\documentclass[letterpaper,10pt,conference]{ieeeconf}

\IEEEoverridecommandlockouts                              
\usepackage{amsmath}
\usepackage{amssymb}

\usepackage{mathpazo}

\usepackage{algorithm}
\usepackage[noend]{algpseudocode}

\usepackage{graphicx}
\usepackage{textcomp}
\usepackage{subfig}

\usepackage{times}
\usepackage{textcomp}

\usepackage{wrapfig}

\usepackage{xargs}                      
\usepackage[dvipsnames]{xcolor}  
\usepackage[colorinlistoftodos,prependcaption,textsize=tiny]{todonotes}
\newcommandx{\unsure}[2][1=]{\todo[linecolor=red,backgroundcolor=red!25,bordercolor=red,#1]{#2}}
\newcommandx{\change}[2][1=]{\todo[linecolor=blue,backgroundcolor=blue!25,bordercolor=blue,#1]{#2}}
\newcommandx{\info}[2][1=]{\todo[linecolor=OliveGreen,backgroundcolor=OliveGreen!25,bordercolor=OliveGreen,#1]{#2}}
\newcommandx{\improvement}[2][1=]{\todo[linecolor=Plum,backgroundcolor=Plum!25,bordercolor=Plum,#1]{#2}}
\newcommandx{\thiswillnotshow}[2][1=]{\todo[disable,#1]{#2}}

\title{\LARGE \bf
Continuous Value Iteration (CVI) Reinforcement Learning  and Imaginary Experience Replay (IER) for learning multi-goal, continuous action and state space controllers
}
\author{Andreas Gerken and Michael Spranger
\\Sony Computer Science Laboratories Inc., Tokyo, Japan}

\begin{document}

\maketitle
\thispagestyle{empty}
\pagestyle{empty}

\begin{abstract}
This paper presents a novel model-free Reinforcement Learning algorithm for learning behavior in continuous action, state, and goal spaces. The algorithm approximates optimal value functions using non-parametric estimators. It is able to efficiently learn to reach multiple arbitrary goals in deterministic and nondeterministic environments. To improve generalization in the goal space, we propose a novel sample augmentation technique. Using these methods, robots learn faster and overall better controllers. We benchmark the proposed algorithms using simulation and a real-world voltage controlled robot that learns to maneuver in a non-observable Cartesian task space.
\end{abstract}

\section{Introduction}
Learning to control one's body is a crucial skill for any embodied agent. A common way of framing the problem of learning to control an agent is Reinforcement Learning (RL). RL poses the problem in terms of actions that an agent can perform, observed states of the world and some reward function that pays out a treat or punishes the agent depending on its performance. The aim of an optimal RL controller is to maximize the collected rewards. Reinforcement Learning has been studied widely and applied to different domains of learning and control.

Suppose we want a robot to learn to control its movements by direct continuous voltage control. Many of the recent prominent RL results \cite{mnih2015human,silver2016mastering} are restricted to \emph{discrete state and discrete action spaces} such as ATARI. Some newer approaches (e.g. \cite{DDPG,wang2016sample,schulman2017proximal}) extend into continuous state and action spaces. However, almost all recent methods rely on huge datasets to perform well (\emph{data efficiency problem}). Such datasets are normally not available for real robots and difficult to obtain. Another problem with current methods is that they work well in learning to reach single goals. For instance, often algorithms learn to reach a particular state (e.g. position in the state space) with an end-effector. However, robots need to learn to achieve different goals. Approaches to obtaining more data and to apply RL to multiple goal tasks can be a physical simulation of the robot \cite{TobinFRSZA17} or \emph{sample augmentation} of existing data \cite{HER}. Especially sample augmentation has seen recent advances but the state-of-the-art methods can only produce a limited number of augmented samples.

\begin{figure}[ht]
  \centering
  \includegraphics[width=0.5\columnwidth]{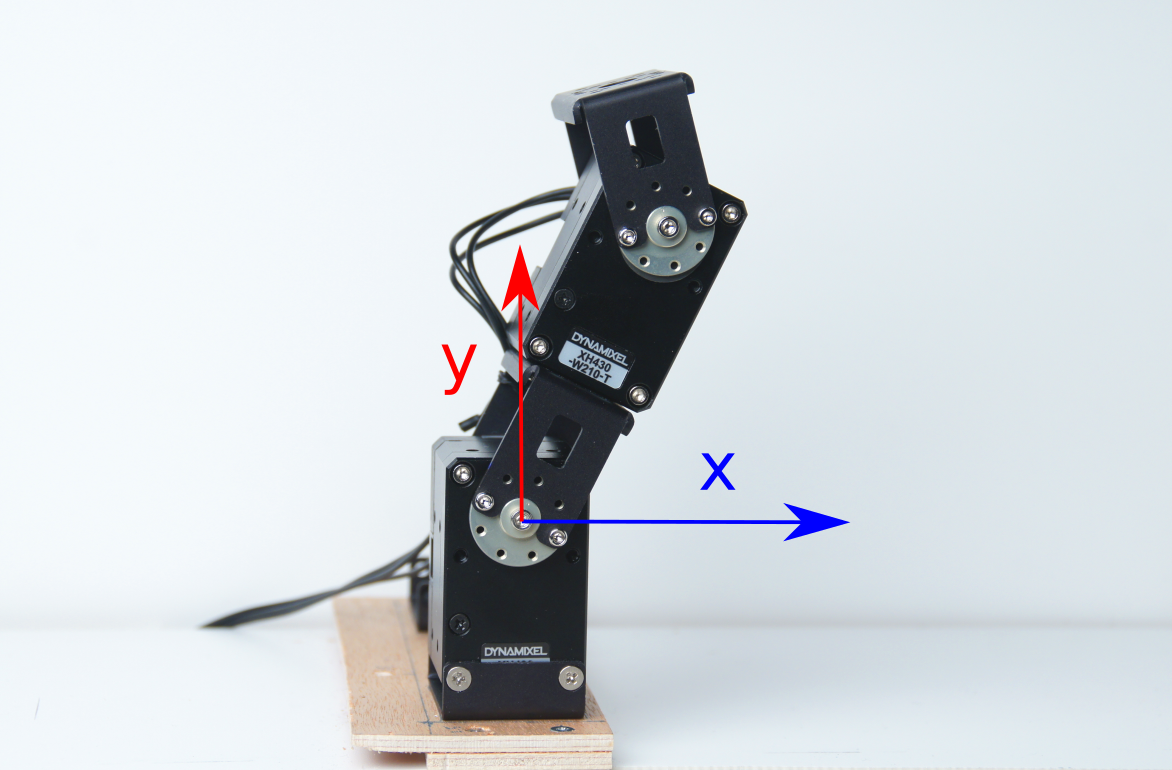}
  \caption{The voltage controlled robot arm consisting of a chain of two Dynamixel XH-430 motors (the coordinate system shows the Cartesian task space of the robot).}
  \label{fig:robot_arm}
\end{figure}

In this paper we try to address these issues by 1) presenting a novel RL approach to learning behavior in continuous state/action spaces with multiple goals called Continuous Value Iteration (CVI) and by 2) presenting a novel data augmentation method called Imaginary Experience Replay (IER). We show that the combination of CVI+IER enables robots to solve tasks efficiently with fewer training examples. We evaluate this in simulation and on a physical voltage controlled robot arm (Figure \ref{fig:robot_arm}).

\section{Continuous state and action MDP with goals}
We frame the problem of learning behavior as a Reinforcement Learning problem. In particular, we assume a standard \emph{continuous action, continuous state Markov Decision Process (MDP)} that describes the world and the (possibly stochastic) effect of actions. We extend the standard RL MDP formulation through a reward function conditioned on goals. We assume an environment $E=(\mathcal{S},\mathcal{A},T,\mathcal{G},R)$ with
\begin{itemize}
\item $\mathcal{S}$ - states $\mathcal{S} \subseteq \mathbb{R}^n$
\item $\mathcal{A}$ - actions $\mathcal{A}\subseteq \mathbb{R}^m$
\item $T$ - transition dynamics $T : \mathcal{S}\times \mathcal{A}\times \mathcal{S} \rightarrow \mathbb{R}$ with $T(s,a,s') := P(s_{t+1}=s' | s_{t}=s,a_t=a)$ because of the Markov property.
\item $\mathcal{G}$ - goals $\mathcal{G} \subseteq \mathbb{R}^k$
\item Reward $R_{g}(s_t,a_t,s_{t+1})\rightarrow \mathbb{R}$ with $s_t,s_{t+1} \in S$, $a \in A$ and $g \in G$
\item Policy $\pi : \mathcal{S} \times\mathcal{G} \rightarrow \mathcal{A}$ for choosing actions
\item Discount factor $\gamma$
\end{itemize}
The robot interacts with the environment by choosing actions and observing states over time. Each trajectory takes the form $\tau=[...,(s_t,a_t,r_t,s_{t+1},g),..]$. The goal for an agent is to choose actions from $\mathcal{A}$ so as to maximize the cumulative discounted reward $R=\sum_{t=0}^{t_{\operatorname{max}}}\gamma^t r_t$. MDPs can be solved by learning a value function $V: \mathcal{S}\times \mathcal{G} \rightarrow \mathbb{R}$ that maps states to a utility value describing the value of the state for achieving the maximum reward for a given goal $g$. $V$ is typically described by the Bellmann Equation \cite{Bellman:1957}
\begin{eqnarray}
\begin{split}
V(s,g) = \operatorname{max}_{a \in A}\int_{s'} T(s,a,s') (R_g(s, a, s') + \\ \gamma V(T(s,a,s'),g))ds'
\end{split}
\label{bellmann}
\end{eqnarray}

which iteratively sets the value of $s$ given $g$ to the maximum over the instant reward and the expected discounted value of the future state when choosing $a$ optimally and acting optimally thereafter. Another function similar to $V$ which is often used in value function approaches is $Q: \mathcal{S}\times\mathcal{G}\times\mathcal{A}\rightarrow\mathbb{R}$ that measures the value of action $a$ given state $s$ and goal $g$.

\section{Related Work}

%
%
%
%
%


Lots of recent work in RL deals with \emph{discrete action and discrete state spaces} using Deep Neural Network such as DQN \cite{DQN}, Double DQN\cite{hasselt2016deep}, and the dueling architecture \cite{wang2016dueling}. In discrete state and action spaces, the integral and the $\operatorname{max}_{a \in A}$ of Equation \ref{bellmann} can be calculated easily. However, these methods are not directly applicable to robotics because they require discretization of state and action spaces. Other algorithms address \emph{continuous state spaces} but \emph{discrete action spaces} environments. Examples of such algorithms are Fitted Value Iteration (FVI) \cite{boyan1995generalization} and Kernel-Based methods \cite{Ormoneit2002KernelBasedRL} which use a model to estimate the distribution of future states and so the integral in Equation \ref{bellmann}. However, such algorithms do not tackle continuous action spaces where the term $\operatorname{max}_{a \in A}$ is not applicable since there is an unlimited amount of actions possible. Similar in problem setting to our work are all RL algorithms dealing with \emph{continuous state and continuous action spaces} such as CACLA \cite{CACLA}, NFQCA \cite{hafner2007neural}, DPG \cite{silver2014DPG} and DDPG \cite{DDPG}. DDPG and variants are actor-critic algorithms that estimate a policy (actor) for choosing actions using the reward signal estimated by a (neural) value function estimator (critic) \cite{konda2000actor}.

In our approach, we solve the MDP without having to iterate over states or actions (discrete state/actions) nor do we explicitly learn a transition model $T$ as model-based RL. Our approach (CVI) is a value function approximation approach and therefore related to DQN and similar algorithms. However, we deal with continuous state/action spaces using simple generalizations in the state and action space through regression. CVI is dealing with continuous state/action spaces which are dominated by actor-critic models. From actor-critic, we differ by not using policy gradients for some actor network. CVI does rely on estimating both $V$ and $Q$ function and faster updates of $V$ vs $Q$ - in that sense, there is some similarity with target network DQN approaches. Against the current trends, CVI (although in principle agnostic to regressor choice) is described and implemented in this paper using a simple non-parametric estimator.

Another line of similarity/difference with recent methods is the problem of \emph{multi goal learning}. Within RL there is some work on multiple goal learning like \cite{precup2001off} or UVFA \cite{UVFA}. Latter proposes to use a single function approximator to estimate the value function of a factored goal and state spaces. Results show that it is possible to generalize over multiple goals if there is a structure in the goal space \cite{Foster2002}. The main differences between our work and UVFA and similar approaches \cite{sutton2011horde} is that they work with discrete state and action spaces. They also do not investigate the impact of sample augmentation. For example, \cite{Yang2017} extends DDPG with continuous action and state spaces to multiple goals but require them to be discrete and limited.

Outside of RL, controllers for robots  without prior knowledge are estimated using \emph{self-exploration}. In motor babbling \cite{demiris2005motor}, random actions are performed and through the resulting observations, a forward model is trained. In goal babbling \cite{rolf2012goal}, goals are set in the task space and during exploration, an inverse model is trained. The placement of the goals can be controlled by intrinsic motivation \cite{baranes2013active} for more efficient exploration of the task space. Similarities of our work and goal babbling are the random placement of goals and the random exploration at the beginning of an experiment. However, CVI does not train an explicit forward or inverse model.

\section{Continuous Value Iteration (CVI)}
We propose \emph{Continuous Value Iteration (CVI)} for learning near optimal controllers in continuous state and continuous action space MDPs. CVI's core is the estimation of the value function $V$ and its subsequent use to approximate $Q$. Past experiences of the robot in the form of trajectories $\tau$ -- i.e. states, actions, rewards, and goals are stored in a \emph{replay buffer} $B$ that consists of tuples $(s_t,a_t,r_t,s_{t+1},g)$. We perform Value-Iteration to propagate the values through the state and goal space. Since both spaces are continuous, we have to achieve generalization using a function approximator. A regressor is used to learn estimates of $V$ and when it has converged, it is used to estimate $Q$. 

In this paper, we use k nearest neighbor (KNN) regression \cite{altman1992introduction} with an Euclidean distance function, however, in principle, other regressors could be used. KNN is a non-parametric method that generalizes well locally. The advantage in this task of KNN over other regressors is its simplicity and that it estimates the values conservatively in regions without training data.

The algorithm (see also Algorithm \ref{algo:CVI}) has 4 Steps (a) action selection and data collection, (b) sample augmentation, (c) $V$ learning, (d) $Q$ learning.

\begin{algorithm}
\caption{Continuous Value Iteration}\label{algo:CVI}
\begin{algorithmic}[1]
\State \textbf{Given:} Function approximators $V(s, g)$, $Q(s, g, a)$
\State \textbf{Given:} Parameters $\nu,\beta,\gamma$ (and $k$ for KNN)
\State Initialize $V_{0}(s,g) = 0$ and $Q_{0}(s,g,a)=0$; $\forall s\in\mathcal{S},\forall g\in\mathcal{G}, \forall a\in\mathcal{A}$
\State Initialize $B=\emptyset$
\Loop
\State // (a) Action selection and data collection
\For{Episode e = 1,E}
  \State Choose $g\in\mathcal{G}$
  \For{Timestep t = 1, N}
  \State Observe $s_t$, choose and execute $a_t$ according to $\pi$, receive reward $r_t$
  \State Save $(s_{t-1}, a_{t-1}, r_t, s_t, g)$ in $B$
  \State Stop when $r_t=1$
  \EndFor
\EndFor

\State \textit{// (b) Sample augmentation}
\State HER, IER

\State \textit{// (c) $V$ Iteration}
\For{Iteration i = 1, I}
\ForAll{$(s, a, r, s', g)$ in $B$}
\State $V_{i+1} \gets [s, g, \operatorname{max}(r, \gamma V_{i}(s',g) , \beta V_{i}(s,g))]$
\EndFor
\State Stop loop when $V$ converges.
\EndFor
\State \textit{// (d) $Q$ learning}
\ForAll{$(s, a, r, s', g)$ in $B$}
\State $Q \gets [s, g, a, V_{I}(s', g)]$
\EndFor
\EndLoop
\end{algorithmic}
\end{algorithm}

\paragraph{Action selection and data collection} In each iteration the robot chooses and performs an action and collects data. This leads to observed trajectories $\tau$ and observed tuples $\tau_{t}$. We store observed tuples in the replay buffer $B$. For training, actions are chosen with an $\epsilon$-greedy action selection policy, where a random action is chosen with the probability of $\epsilon$ and the best action is chosen otherwise. This helps the algorithm to exploit existing knowledge during exploration. When using and validating the policy, a greedy policy following the most valuable action is used.

\paragraph{Sample augmentation} Data in the replay buffer $B$ can be augmented using various techniques. We employ three strategies: 1) None: the buffer stays as is, 2) Hindsight experience replay (HER), 3) Imaginary Experience Replay (IER). HER and IER add $\tau_t$ tuples to the replay buffer $B$ using different strategies. More technical detail will follow in Sections \ref{sec:HER} and \ref{sec:IER}.

\begin{figure}
\centering
  \includegraphics[width=0.2\textwidth]{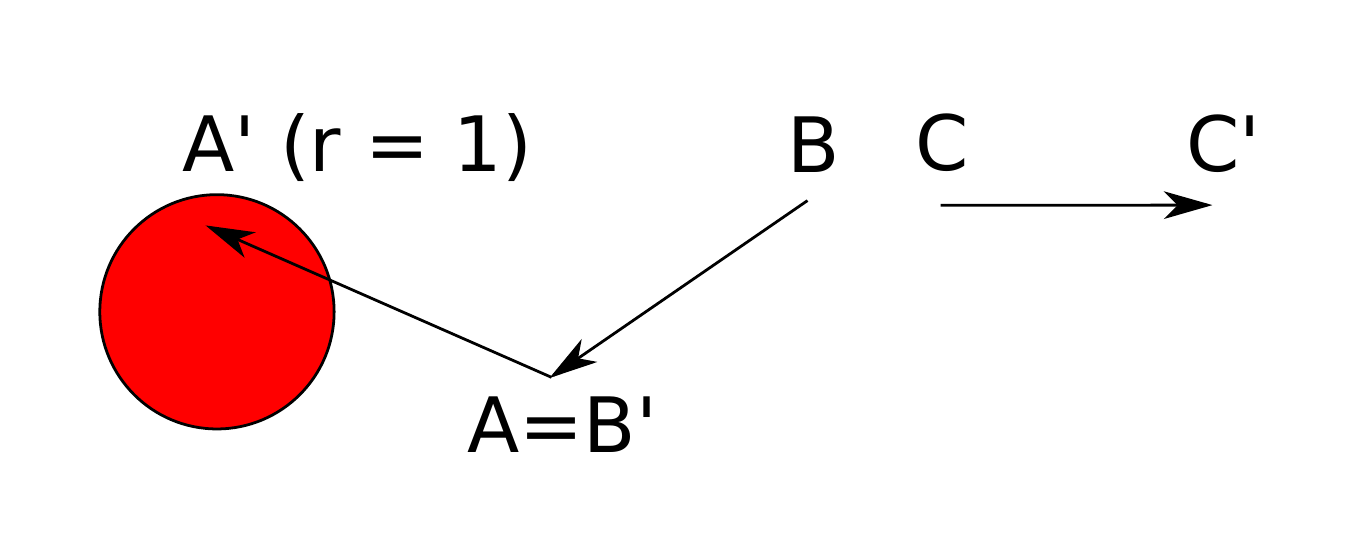}
  \caption{Two trajectories with three transitions in the point environment and a red goal region with a reward of 1}
  \label{fig:show_cvi}
\end{figure}

\paragraph{$V$ iteration} We compute new estimates of $V$ using the replay buffer $B$ as input. The algorithm iterates $I$ times over the entire replay buffer $B$ and computes training data for the KNN regressor.

\begin{eqnarray}
\label{eq:v_update}
V^{i+1}\leftarrow [s,g,\operatorname{max}(r,\gamma V^i(s', g),\beta V^i(s, g)]
\end{eqnarray}

Training data for $s,g$ pairs is computed by taking the maximum value of the following three sources. We explain the sources referring to Figure \ref{fig:show_cvi}.
\begin{itemize}
    \item Reward $r$: The reward is an immediate source of value if a goal state was reached. For examples, this term is chosen for point $A$ (Figure \ref{fig:show_cvi}), because the action directly leads to a reward.
    \item Discounted predicted value $\gamma V^i(s',g)$: This value is the same as the Bellmann backup except here it is predicted by the regressor. This term spreads value along successful observed trajectories. For example, this term is chosen for point B (Figure \ref{fig:show_cvi}), when the future state ($B'=A$) has a value. The hyper-parameter $\gamma$ is the typical discount factor in MDPs. We use $\gamma=0.99$.
    \item The estimated value of the current state $\beta V^i(s, g)$: This allows the algorithm to spread values through the state and goal space. The term implicitly replaces the search for the best action in Equation \ref{bellmann} ($\operatorname{max}_{a \in A}$). For example, this term is chosen for point $C$ (Figure \ref{fig:show_cvi}), if the neighboring state $B$ has a high value. The cooling factor $\beta$ counteracts overestimations from previous $V$. With an $\epsilon$-greedy agent, areas with an overestimated value are explored more in future episodes, so that errors are quickly fixed.
\end{itemize}

\section{Sample Augmentation:}
Robots can only gain limited experience from the environment constrained by the real-time movements and the sampling rates of sensors and actuators. To enhance the training data it can be enhanced by additional samples, which can be inferred without additional knowledge like physics simulations. In this paper, we propose a new approach (IER) and compare it to an existing approach (HER).
\subsection{Hindsight Experience Replay (HER)}
\label{sec:HER}
HER \cite{HER} is a recent example of sample augmentation where experiences (samples) are added to the buffer for additional goals. HER assumes that states can be goals and therefore states later in a trajectory/episode can be assigned as goals to samples earlier in the trajectory thereby creating new samples. The new data is added to the replay buffer $B$. The newly created samples are limited to have goals which are previously seen states. The maximum amount of samples HER can create for a trajectory of length $n$ is $\frac{n (n+1)}{2}$. Importantly, HER assumes that goal and state space are equal, i.e. that goals are states in the state space. Therefore, HER cannot be applied to domains, where the goal and state spaces are separate.

\subsection{Imaginary Experience Replay (IER)}
\label{sec:IER}
We address the main limitations of HER in this paper by proposing Imaginary Experience Replay (IER). IER is able to 1) produce infinite amounts of samples from finite experiences $B$ and 2) deal with separate goal and state space domains. IER does this by extending experienced transitions with \textit{imaginary} goals. The algorithm (see also Algorithm \ref{algo:IER}) samples any number $S$ of additional samples from $B$. For all additional samples, a new random goal $\hat{g} \in G$ is sampled and the reward is changed according to $R_{\hat{g}}(s_{t},a_{t},s_{t+1})$. IER applied to CVI helps in spreading discounted rewards through the $V$ (and $Q$) landscape and therefore aids generalizations across different goals. Samples created by IER can serve as the glue between experienced trajectories with different goals.

\begin{algorithm}
\caption{Imaginary Experience Replay (IER)}\label{algo:IER}
\begin{algorithmic}[1]
\State \textbf{Given:} \\
Replay buffer $B$ with transitions ($s, a, r, s', g$) \\
Goal space $G$ and a way to sample from $G$\\
Reward function $R_g(s,a,s')$
\For{Sample s = 1, S}
\State Sample imaginary goal $\hat{g}$ from G (using any distribution, we use uniform)
\State Sample transition $(s, a, r, s', g)$ from replay buffer $B$
\State Store $(s, a, R_{\hat{g}}(s,a,s'), s',\hat{g})$ in replay buffer $B$
\EndFor
\end{algorithmic}
\end{algorithm}

\section{Experiment I: Simulated point environment}
We validate CVI and IER using two types of environments. The first environment is a simulation of a moving two-dimensional point on a plane. We use the simulation to explore the impact of design choices and compare our method with existing state-of-the-art methods.

\subsection{Experimental Setup}
\label{experimental_setup}
\paragraph{State space $\mathcal{S}\subseteq \mathbb{R}^2$} with $s=[x,y]$ agent position.
\paragraph{Actions $\mathcal{A}\subseteq \mathbb{R}^2$} with $a=[dx, dy]$ agent velocity.
\paragraph{Transition function $T(s,a,s') := P(s_{t+1}=s' | s_{t}=s,a_t=a)$}
\paragraph{Goals $\mathcal{G}=\mathcal{S}$} with a fixed margin $w$ that determines if a goal has been reached. We study two types of experiments:
\begin{itemize}
\item \emph{one goal} - goal is the same for the duration of a particular experiment (training and evaluation).
\item \emph{random goal} - new goals are chosen randomly from $\mathcal{G}$ when a goal has been reached.
\end{itemize}

\paragraph{Reward function $R_g$} The reward function is binary: a state $s$ is considered a goal state iff $|s-g| < w$ and the reward for goal states is one. For all other states reward is zero.
\begin{eqnarray}
R_g(s,a,s')=\begin{cases}
1, \text{iff } |s'-g| < w\\
0, otherwise
\end{cases}
\end{eqnarray}

\paragraph{Training} For each training episode, the agent's position is set to a random location. In each episode, a maximum of 200 transitions (timesteps) can be performed. The agent position is set randomly and the agent gets 30 timesteps to reach the goal. The agent's position is reset randomly if the agent reaches the goal or 30 timesteps are up. If 200 timesteps are up, the training episode is finished. The agent can reach a variable number of goals within a training iteration.

\paragraph{Evaluation}
\label{evaluation} We evaluate the controller after each training episode. The maximum length of an evaluation episode is 2000 timesteps. The agent is randomly set (and in \emph{random goal} tasks a goal is randomly chosen). For each trial the agent then has a maximum of 30 timesteps to reach the goal. The agent's position is reset when reaching the goal or after 30 timesteps. The performance of the controller is quantified by comparing the agent's trajectory with the analytically calculated optimal trajectory. Suppose the agent took $m$ steps to reach the goal with $\operatorname{opt}$ being the shortest possible number of steps to reach the goal, then the \emph{optimal control score} is
\begin{eqnarray}
\operatorname{score}(m)=
\begin{cases}
1-\frac{m-\operatorname{opt}}{30-\operatorname{opt}}, \text{iff agent reaches goal}\\
0, otherwise
\end{cases}
\end{eqnarray}
with $\operatorname{score}(m)=1$ if the agent takes as many timesteps to the goal as the optimal trajectory would and $0$ if the goal is never reached. The average score of all trials is the final score.

\paragraph{Benchmarking}
We evaluated various systems to understand the performance of CVI and the effect of sample augmentation. We benchmark our own sample augmentation technique (IER) against the state of the art (HER). We execute the same experiment 10 times for each of the following systems.

\begin{itemize}
\item\emph{CVI:} vanilla CVI without augmentation
\item\emph{CVI+HER:} CVI with HER sample augmentation
\item\emph{CVI+IER:} CVI with IER sample augmentation. The number of samples added is equal to the number of samples HER can produce for the given replay buffer $B$.
\item\emph{CVI+IER 3X:} CVI with IER sample augmentation. The number of samples is the same that HER can produce times 3.
\item\emph{CVI+IER 10X:} CVI with IER sample augmentation. The max number of samples is the same that HER can produce times 10.
\end{itemize}

\subsection{Results I: CVI learns to solve continuous state and action space tasks}
\begin{figure}
  \centering
  \includegraphics[width=0.46\textwidth]{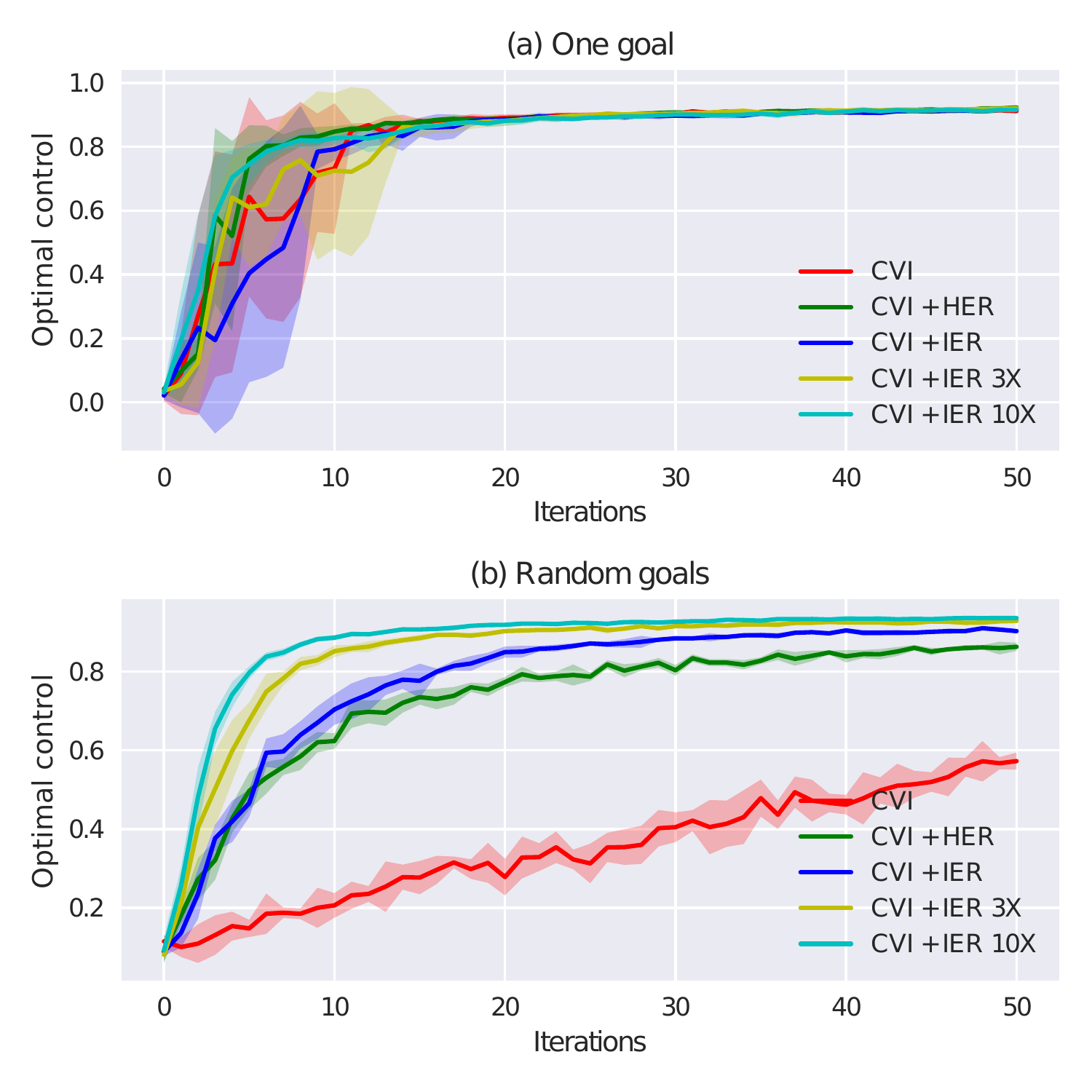}
  \caption{Performance of CVI with various sample augmentation strategies in \emph{one goal} (a) and \emph{random goal} (b) tasks in the point environment. The y-axis shows, how close the agent is to optimal control (see Section \ref{experimental_setup}). For each configuration, 10 independent runs were performed and  the averages with their bands of $\pm \sigma$ are shown. The KNNs use k=5 neighbors.}
  \label{fig:point_env_no_obstacles_cvi+augmentation}
\end{figure}

We first evaluate CVI's ability to solve \emph{one goal} tasks. Figure \ref{fig:point_env_no_obstacles_cvi+augmentation}a shows that CVI (even without any sample augmentation strategies) is able to quickly solve \emph{one goal} tasks in the point environment. We see convergence reliably after 10 training iterations (2000 timesteps). Figure \ref{fig:point_env_no_obstacles_cvi+augmentation}a also compares the performance of different sample augmentation strategies. It is clear that for the \emph{one goal} tasks in a point environment sample augmentation is not necessary to achieve a good performance.

\begin{figure}[ht]
  \centering
  \subfloat[Predicted values $it=1$]{{\includegraphics[width=0.2\textwidth]{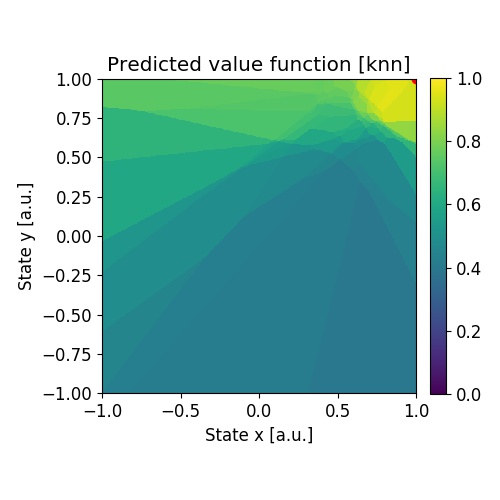} }}%
  \qquad
  \subfloat[Predicted values $it=12$]{{\includegraphics[width=0.2\textwidth]{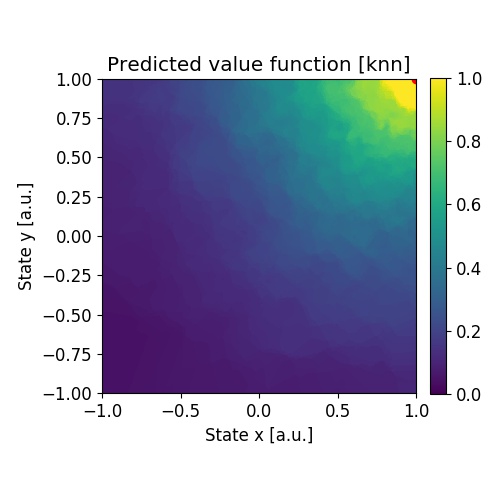} }}%
  \qquad
  \subfloat[Predicted value $it=100$]{{\includegraphics[width=0.2\textwidth]{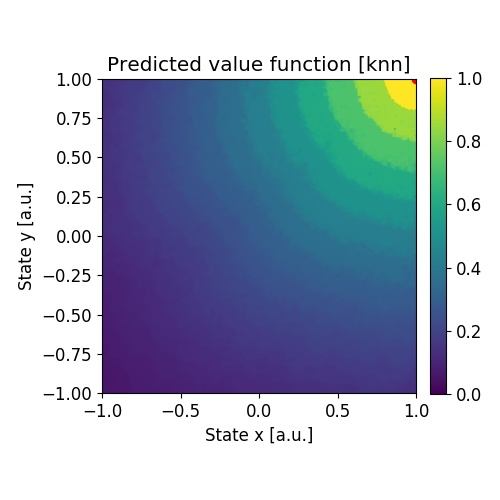} }}%
  \qquad
  \subfloat[Actual Value of states]{{\includegraphics[width=0.2\textwidth]{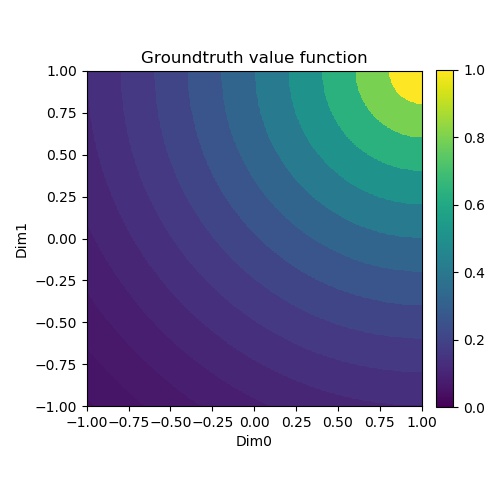} }}%
  \caption{Value function in the point environment with fixed goal at $g=(1,1)$ ($d_{max}=0.2$ and $w=0.2$). (a - c) prediction over time with CVI ($\gamma = 0.85$ and $\beta = 0.99$); (d) analytically calculated.}
  \label{fig:predicted_real_value}
\end{figure}

Figure \ref{fig:predicted_real_value} shows the learned reward value $V$ for states in the state space (\emph{one goal}). We measure this by sampling a collection of states $s$ and plotting their $V_i(s, g)$ (with fixed $g$ and $i\in[1,12,100]$). We can see that after few iterations ($12$) (Figure \ref{fig:predicted_real_value}b) the value function approximates the true underlying discounted reward value landscape (optimal $V^*$) shown in Figure \ref{fig:predicted_real_value}d. This explains why the agent is able to collect rewards quickly. Notice that basically the task is solved after 10-12 iterations. After that, the $V$ estimates become even more accurate, however, this is not strictly necessary for good task performance. All it takes is for the landscape of $V(s,g)$ to have similar local derivatives as $V^*(s)$.

We did similar experiments in an environment with an obstacle (a virtual wall). Results are omitted here due to space limitations, but CVI solved that environment equally quickly.

\subsection{Results II: CVI + IER learn to solve tasks with different goals}
We evaluated CVI's ability to solve \emph{random goal} tasks. Figure \ref{fig:point_env_no_obstacles_cvi+augmentation}b shows that CVI is able to solve \emph{random goal} tasks in the point environment. However, here sample augmentation strategies clearly improve the performance of the system significantly. CVI alone is learning to solve the task, however, adding sample augmentation strategies aids early convergence. We see convergence starting from 10 training iterations (2000 samples) in the best case. This is comparable to the \emph{one goal} environment with almost no loss in learning speed.

Figure \ref{fig:point_env_no_obstacles_cvi+augmentation}b also compares the performance of different sample augmentation strategies. We can see that IER outperforms HER, even with the same amount of additional samples. The main reason for this is that IER allows for more generalization in the goal space through much more samples with varying goals. Importantly, adding more goals does not significantly slow learning when using IER. Convergence is slower, but adding infinitely many goals to the task has almost negligible effects compared to the increased difficulty.

\subsection{Results III: CVI vs DDPG}
We compared CVI with the current continuous action/state space state-of-the-art algorithm DDPG. DDPG uses a replay buffer to directly optimize a policy. We used the DDPG implementation of OpenAI\footnote{https://github.com/openai/baselines/tree/master/baselines/ddpg} on the point environment and optimized DDPG hyper-parameters with full grid search. We show the results of the best configuration.

\begin{figure}
  \centering
  \includegraphics[width=0.46\textwidth]{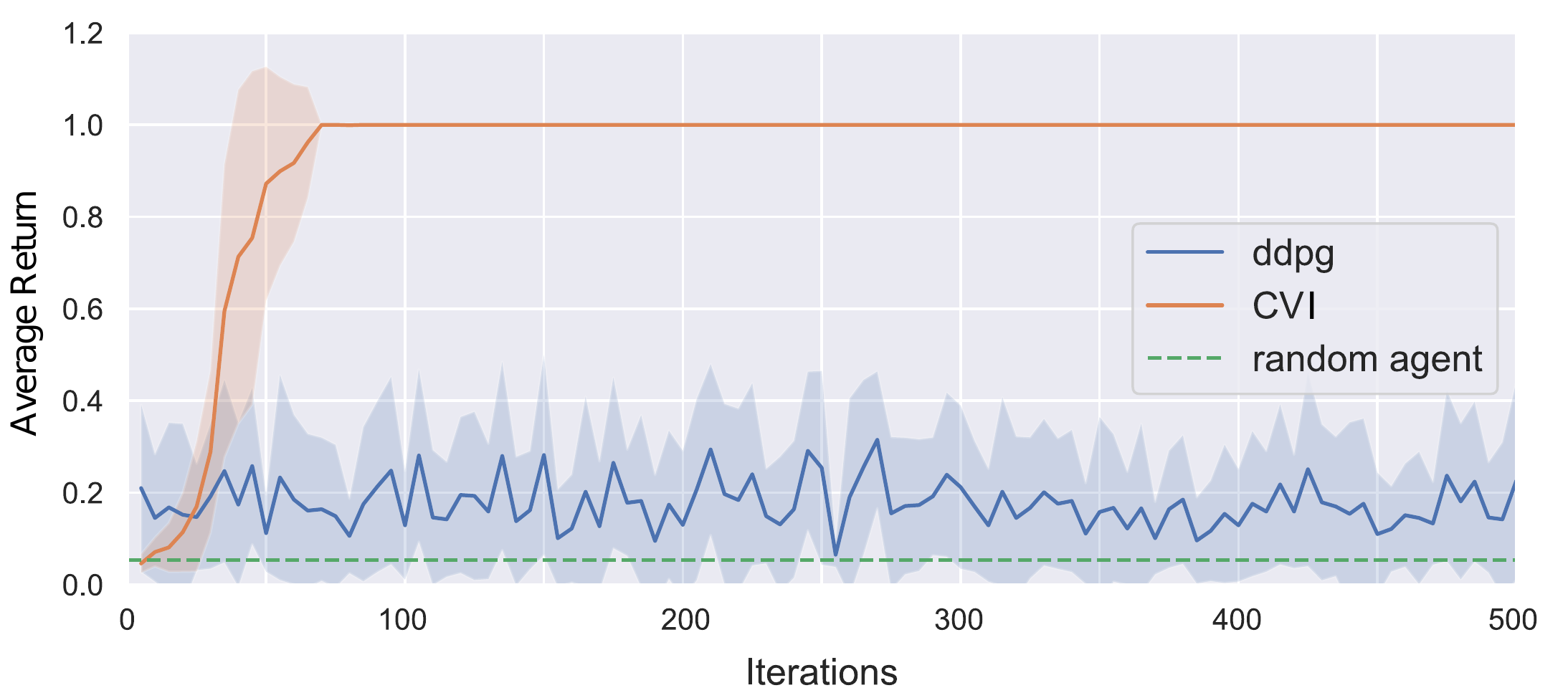}
  \caption{Comparison of CVI and DDPG in the point environment ($d_{max}=0.05$, $w=0.1$). The KNNs use k=5. In the x-axis, algorithm iterations are shown where one iteration includes 200 state transitions in the environment and in the y-axis, the average return per test episode with $\pm \sigma$ is shown where 1 means that the goal is reached in all episodes.}
  \label{fig:cvi_ddpg}
\end{figure}

Figure \ref{fig:cvi_ddpg} shows that DDPG does not solve the environment whereas CVI quickly learns a near optimal controller. Further experiments showed that the main reason for DDPGs failure is the sparse reward structure of the task. In contrast to CVI, DDPG does not seem to be able to handle very sparse rewards. Notice that these experiments were done in the \emph{one goal} point environment. Since DDPG was unable to solve this task we did not extend evaluation to random goals. Also, we omitted sample augmentation since HER, IER and other goal sample augmentation techniques have no effect in this setting.

\section{Experiment II: Robot arm}
We use a robot attached to a table to show that CVI can learn continuous controllers in the real-world. The robot consists of a kinematic chain of Dynamixel motors (see Figure \ref{fig:robot_arm}, page \pageref{fig:robot_arm}). We use this experimental setup to validate CVI in two ways. First, we are interested in direct voltage motor control in the real-world. Second, we evaluate whether the system learns to reach coordinates in an absolute Cartesian space (task space) without directly providing observable access to the Cartesian space. In other words, the robot's state representation does not include the task space. The robot has to learn to control task space through sparse reward feedback only.

\subsection{Experimental Setup}
\paragraph{State space $\mathcal{S}$} The state space of the robot is the joint space and joint velocity of the two actuators $s=[r_1,r_2,\dot{r_1},\dot{r_2}]$. The state is directly measured by hardware sensors.

\paragraph{Actions $\mathcal{A}$} The motors are directly controlled by setting voltages $a=[v_1,v_2]$. The Dynamixel uses PWM to control average supply voltage to the motor based on $a$ and thereby controls the torque applied to the motor. The Motors are backdrivable and the robot is not statically stable. The robot just falls down, when the voltage is zero. Similarly, the robot requires some control signal to overcome joint friction.

\paragraph{Goal space $\mathcal{G}$} Goals are positions in the Cartesian space $(x,y)$. The Cartesian space is centered in the middle of the first joint and fixed along the axis of the table (see Figure \ref{fig:robot_arm}). I.e. the goal space is relative to the table and does not rotate with the robot. Notice that state and goal space are completely separate spaces. Goals are not part of the state space and only indirectly accessible to the robot via sparse reward signals.

\paragraph{Reward $R$} The reward function is the same as in the point environment. We measure the position of the end effector in goal space using internal readouts of the motor rotations and prior knowledge of the forward model to calculate the absolute position in space with respect to the Cartesian coordinate system. This is used exclusively in the reward function. The robot does not have access to the forward model.

\paragraph{Training} We train the system for 60 minutes ($\sim$ 20k experienced state transitions at 5 fps). The robot starts in a random position. A goal is chosen and the robot has 100 timesteps (20 sec) to reach the goal. If it reaches the goal or time is up, the goal is reset randomly. Notice that in contrast to the point environment, the robot position is never reset. Also, this is effectively a random goal environment.

\paragraph{Evaluation} We test the system for 10 minutes (3000 timesteps at 5fps) with a similar setup as in training. We choose a random goal and the robot has a maximum of 100 timesteps (20 seconds) to reach that goal. The goal is reset if the robot reaches the goal or time is up. We measure how many rewards the robot was able to collect in 10 minutes (cumulative reward).

The experiments are repeated 10 times each to obtain statistical data.

\begin{figure}[ht]
  \centering
  \includegraphics[width=0.5\textwidth]{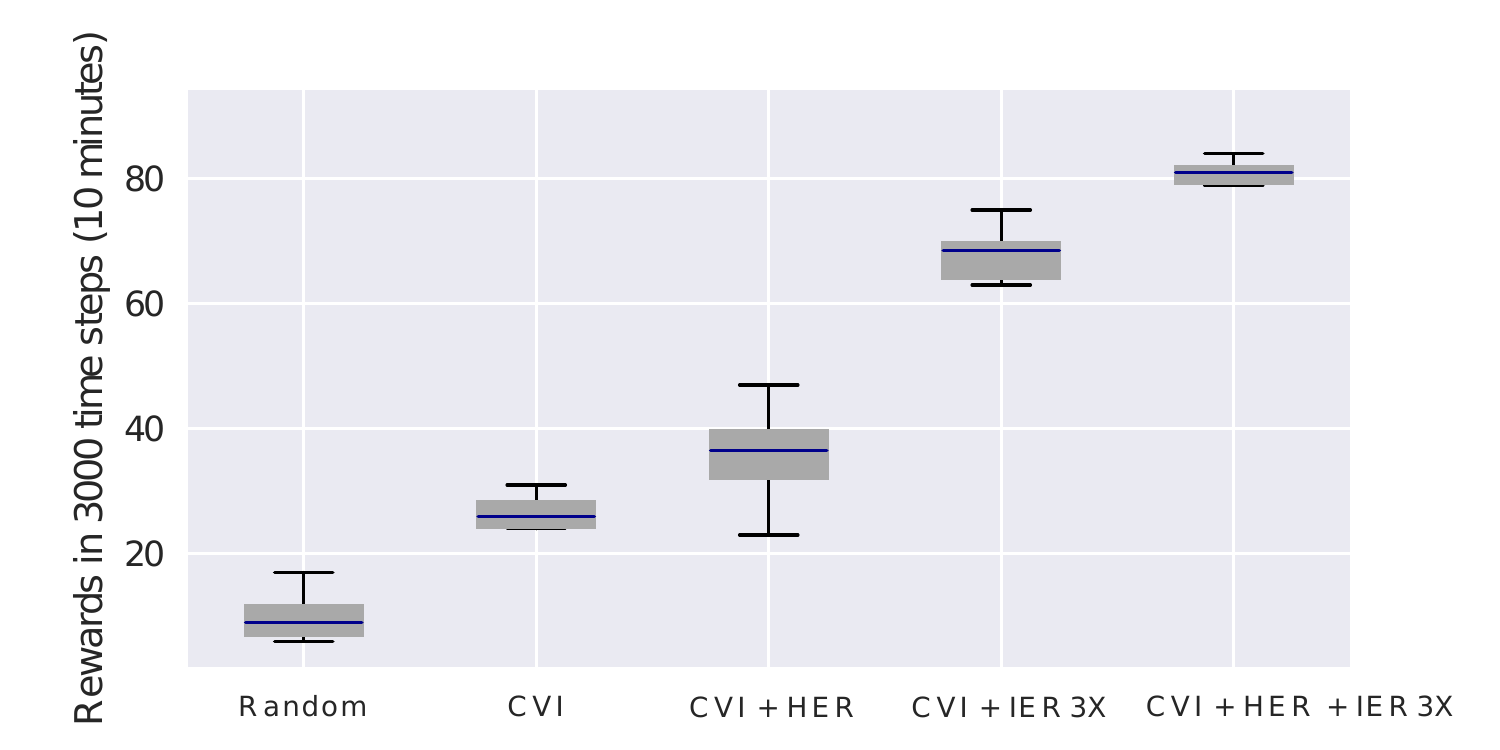}
  \caption{Average cumulative reward in 3000 timesteps (10 minutes) of test data on the robot arm for various algorithms}
  \label{fig:success_arm}
\end{figure}

\subsection{Results and Discussion}

Figure \ref{fig:success_arm} shows the performance of various strategies of CVI with and without augmentation on the real robot with voltage control action space and Cartesian goal space. CVI significantly outperforms the random control policy. The results also show that sample augmentation significantly helps CVI. HER gives some improvement but maybe not statistically significant. However, IER and especially IER+HER make the system perform very well. If we check the performance of CVI+HER+IER we can see that in 10 minutes it reaches an average of 80 goals. That directly translates into reaching a new goal on average every 7.5 seconds.

In our view, the performance of the system is remarkable. CVI (HER+IER) learns to perform actions directly in the real-world. Here, CVI does not use stable position control but directly manipulates the voltage of the motors and has to deal with physical effects such as gravity, friction without receiving lots of sensory information (only joint position and velocity). Moreover, the robot learns to control a task space to which it has no direct access. It can only indirectly access information about this space via the reward function. Reward though is binary and is only experienced when reaching a goal. This shows that CVI can efficiently solve continuous action and state space RL problems with sparse rewards.

\section{Conclusion}

We have presented two novel methods for Reinforcement Learning in continuous state, goal and action spaces. Firstly, Continuous Value Iteration (CVI) enables the efficient estimation of utility functions $V$ and $Q$. We see that these methods generalize well in continuous state, action, and goal spaces. Second, Imaginary Experience Replay (IER) significantly enhances the performance of CVI by adding potentially unlimited amounts of samples for better generalization. We have shown in two environments that the proposed methods perform well. Importantly, CVI+IER enable a voltage controlled real robot to quickly learn to move in the real-world without explicitly learning forward or inverse models.

\bibliography{bibliography}{}
\bibliographystyle{ieeetr}
\end{document}